%% file: main.tex
\documentclass{article}
\usepackage{ijcai22}

\usepackage{times}

\usepackage{soul}
\usepackage{url}
\usepackage[hidelinks]{hyperref}
\usepackage[utf8]{inputenc}
\usepackage[small]{caption}
\usepackage{graphicx}
\usepackage{amsmath}
\usepackage{booktabs}
\usepackage{multicol}
\usepackage{multirow}
\usepackage{arydshln}
\usepackage{subcaption}
\usepackage{bbm}
\usepackage{pgfplots}
\usepackage{hyperref}
\usepackage{tikz}
\usepackage{pgf}
\newcommand{\ra}[1]{\renewcommand{\arraystretch}{#1}}

\title{Raw Produce Quality Detection with Shifted Window Self-Attention}

\author{
Oh Joon Kwon\footnote{Equal contribution.}\and
Byungsoo Kim$^*$\and
Youngduck Choi
\affiliations
Togi Inc.\\
\emails
\{oj.kwon, bs.kim, yd.choi\}@togi.co,
}

\begin{document}

\maketitle

\input{1_abstract}

\input{2_introduction}

\input{3_related_works}

\input{4_models}

\input{5_datasets}

\input{6_experimental_results}

\input{7_conclusion}

\newpage
\clearpage
\bibliographystyle{named}
\bibliography{ref}

\end{document}

%% file: 1_abstract.tex
\begin{abstract}
Global food insecurity is expected to worsen in the coming decades with the accelerated rate of climate change and the rapidly increasing population.
In this vein, it is important to remove inefficiencies at every level of food production. 
The recent advances in deep learning can help reduce such inefficiencies, yet their application has not yet become mainstream throughout the industry, inducing economic costs at a massive scale. 
To this point, modern techniques such as CNNs (Convolutional Neural Networks) have been applied to RPQD (Raw Produce Quality Detection) tasks.
On the other hand, Transformer's successful debut in the vision among other modalities led us to expect a better performance with these Transformer-based models in RPQD.
In this work, we exclusively investigate the recent state-of-the-art Swin (Shifted Windows) Transformer which computes self-attention in both intra- and inter-window fashion.
We compare Swin Transformer against CNN models on four RPQD image datasets, each containing different kinds of raw produce: fruits and vegetables, fish, pork, and beef.
We observe that Swin Transformer not only achieves better or competitive performance but also is data- and compute-efficient, making it ideal for actual deployment in real-world setting.
To the best of our knowledge, this is the first large-scale empirical study on RPQD task, which we hope will gain more attention in future works.
\end{abstract}

%% file: 2_introduction.tex
\input{figure_tex/model_swin}

\section{Introduction}
In 2020, it was projected that 768 million people faced hunger, which was an approximately 18.5\% increase in number from the year before.
Meanwhile, the world population is expected to reach 9.8 billion in 2050~\cite{unfood}.
We now face the threat of global food insecurity in the coming decades from the rapidly rising demand, coupled with diminishing natural resources and accelerated climate change. 
However, guaranteeing the safety of food in terms of freshness and nutrients at a massive scale while meeting the ever-increasing demand for productivity is a challenging task for sustaining a functional, healthy population. 
Hence, reducing any economic inefficiency from the food manufacturing and inspection procedures is imperative in the food industry.

To this term, quick and accurate automated measure for determining food attributes is in practical demand.
While several modern techniques for collecting food inspection data suggest using sophisticated sensory information such as spectral imaging \cite{al2018detection} and chemical composition \cite{di2017fusion}, analyzing natural images can be as effective and bring drastic change in improving production efficiency. 
The recent success of computer vision proved the possibility of widespread adoption of artificial intelligence in the industry. 
For instance, recent works in agriculture and food focused on utilizing vision models equipped with CNNs (Convolutional Neural Networks) for food production and inspection \cite{sanga2020mobile,masood2020early,valdez2020apple,rasmussen2020evaluation}. 

Transformer architecture, first proposed in \cite{transformer} for neural machine translation, has shown remarkable success in various modalities including natural language processing \cite{bert} and audio processing \cite{audiot}. 
More recently, ViT (Vision Transformer) \cite{vit} achieved the state-of-the-art performance in image classification task without relying on the hard-wired connection induced by convolutional kernels.
Since then, its variants have continued to set new records in numerous vision tasks such as segmentation and object detection.

In this paper, we compare the conventional CNNs with the state-of-the-art Transformer-based vision model on four large scale RPQD (Raw Produce Quality Detection) datasets, fruits and vegetables, fish, pork, and beef. 
In particular, we explore the possibility of applying the current state-of-the-art model called Swin (Shifted Windows) Transformer to RPQD task.
We show that it achieves better overall test performance than the CNN models, achieving 0.3\%, 26.6\%, and 6.4\% higher metrics in the fruits and vegetables, pork, and fish datasets and 0.1\% less in the beef dataset than the best performing CNN model.
We also observe that Swin Transformer is data-efficient; the performance gap between CNN and Swin Transformer widens as the number of training samples decreases. In particular, Swin Transformer performs 75.2\% better than the best performing CNN when only 1/16-th of original training samples are used. 
Moreover, we observe that Swin Transformer has good accuracy-computation trade-off when compared to CNNs, making it ideal for actual application at deployment.




%% file: figure_tex/model_swin.tex
\begin{figure*}[t]
    \centering
    \includegraphics[width=0.8\textwidth]{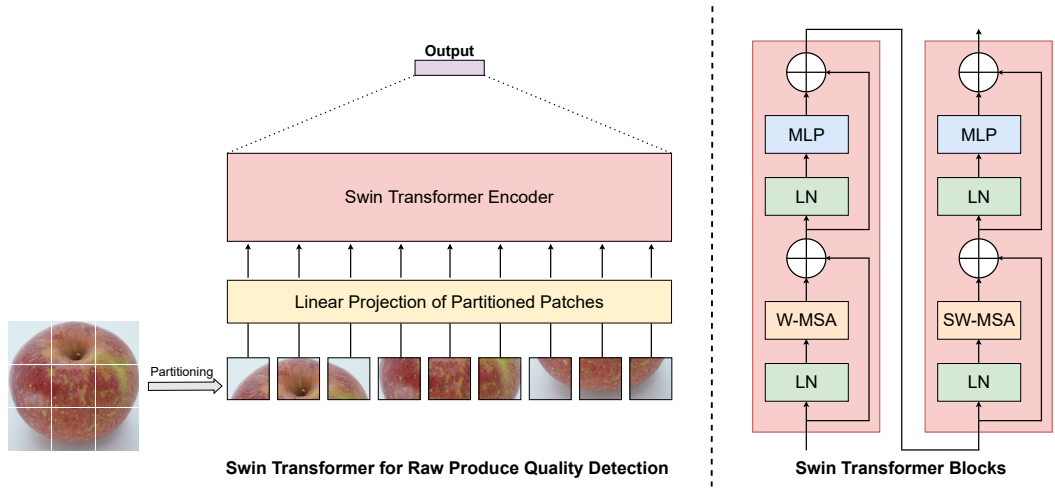}
    \caption{The overview of Swin Transformer. Swin Transformer is a stack of two successive Swin Transformer blocks. W-MSA (Window-based Multi-head Self-Attention) and SW-MSA (Shifted Window-based Multi-head Self-Attention) are distinctive layers in each block.}
    \label{fig:model_swin}
\end{figure*}

%% file: 3_related_works.tex
\section{Related Works}

\subsection{Deep Learning in RPQD}

\cite{rubanga2020deep} defined the three stages of severity of Tuta absoluta pest’s effect in tomato plants, and proposed CNN models (VGG, ResNet, and Inception-V3) to classify each tomato leaf image to one of the severity levels.
\cite{valdez2020apple} compared SSD against YOLOv3 in identifying whether a given image presents a healthy or damaged apple, and detecting the defective part in the image.
The similar approach was taken in \cite{masood2020early} where Mask R-CNN based rice crop disease diagnosis model classifies rice crop image to one of healthy or diseased, and localizes diseased portion.
\cite{rasmussen2020evaluation} formulated the problem of kernel fragment recognition as object detection.
Various CNN models were investigated including ResNet, Inception-V2, MobileNetV1, and MobileNetV2 for backbones, SSD, Faster R-CNN, and Mask R-CNN for bounding box detectors to measure trade-off between model complexity, accuracy, and speed.
\cite{sanga2020mobile} developed FUSI Scanner, a mobile application for banana disease detection. The application deployed Inception-V3 to classify each banana leaf image to one of three diseased states. 

\subsection{Datasets for RPQD}
Although machine learning is a promising tool for food and agriculture, its wide-spread adoption is limited by the lack of publicly available datasets. Most datasets for RPQD were collected in-house and not open to the public \cite{rubanga2020deep,valdez2020apple,masood2020early,rasmussen2020evaluation,sanga2020mobile}, and only a handful are publicly available~\cite{softwaremill_2020,hughes2015open}.
Recently, the ministry of Science and ICT (Information and Communication Technology) of South Korea published AI Hub \footnote{\url{https://aihub.or.kr}}, a data repository for large scale publicly available datasets in various domains including computer vision, natural language processing, agro-livestock, autonomous driving, and healthcare.
In particular, datasets in agro-livestock domain contain diverse images of fruits and vegetables, crops, fish, and meat, and annotations, such as grade, disease, and pest damage.

\subsection{Hierarchical Vision Transformers}

Subsequent variants of ViT (Vision Transformer) focused on addressing the lack of inductive bias by directly integrating CNN into the model~\cite{coatnet} or distilling from a teacher CNN~\cite{deit} to achieve data efficiency at pre-training.
Other approaches used hierarchical structures to capture the variations in the scale of visual objects.
Pyramid ViT~\cite{pvit} proposed a shrinking pyramid and spatially reducing attention. Segformer~\cite{segformer} improved on this with overlapping patch embedding and depth-wise convolution. 
The most recent state-of-the-art model is Swin (Shifted Windows) Transformer~\cite{swint} that limits the self-attention computation to local windows while allowing for cross-window connections by shifting them along the hierarchy.

%% file: 4_models.tex
\input{figure_tex/size_dist}
\input{figure_tex/fusion}

\section{Model}
\subsection{Swin Transformer}
Unlike ViT with uniformly scaled patches, Swin Transformer takes a hierarchical, multi-scale approach to capture the varying scales naturally occurring in images.
Swin Transformer uses small patches ($< 10$ pixels compared to $\geq 16$ in ViT), which are merged together to form windows. 
The conventional self-attention block of Transformer is replaced with a modified muliti-headed self-attention that attends only within each non-overlapping window to circumvent the quadratic computation cost of self-attention on the entire image patches.
Since this lacks connections across windows, the subsequent layers compute self-attention on shifted windows, allowing patches within a window to draw connections to those in other windows. The overview of the model is given in Figure~\ref{fig:model_swin}.
Swin Transformer achieves state-of-the-art and competitive performance in numerous vision tasks that have previously been occupied by CNN, becoming the standard model across vision tasks \cite{swint}.

\subsection{Multi-view Fusion for Elongated Images}
As shown in Figure~\ref{fig:size_dist}, the pork dataset contains elongated image samples (i.e. larger in either height or width dimension than the other).
Applying naive downsampling according to the image size in such cases may lead to significant loss of information.
Moreover, the region of interest (e.g. fatty part) that contributes to the grade of the specimen is often concentrated in a particular area. 
Generating training samples from random crops of such image will degrade the performance as the model will not be able to look at the region of interest from time to time.
In order to integrate features from different parts of the image without distortions, we approach the pork grade classification as a multi-view representation learning.
We create different views by taking top, middle, and bottom sections of each sample.
Following \cite{seeland2021multi}, we fuse the feature vectors of the different views by concatenation as shown in Figure~\ref{fig:fusion}. 
We also compare the performance of model with multi-view fusion and model simply trained on random crops of an image.


%% file: figure_tex/size_dist.tex
\begin{figure}[t]
    \centering
    \includegraphics[width=\columnwidth]{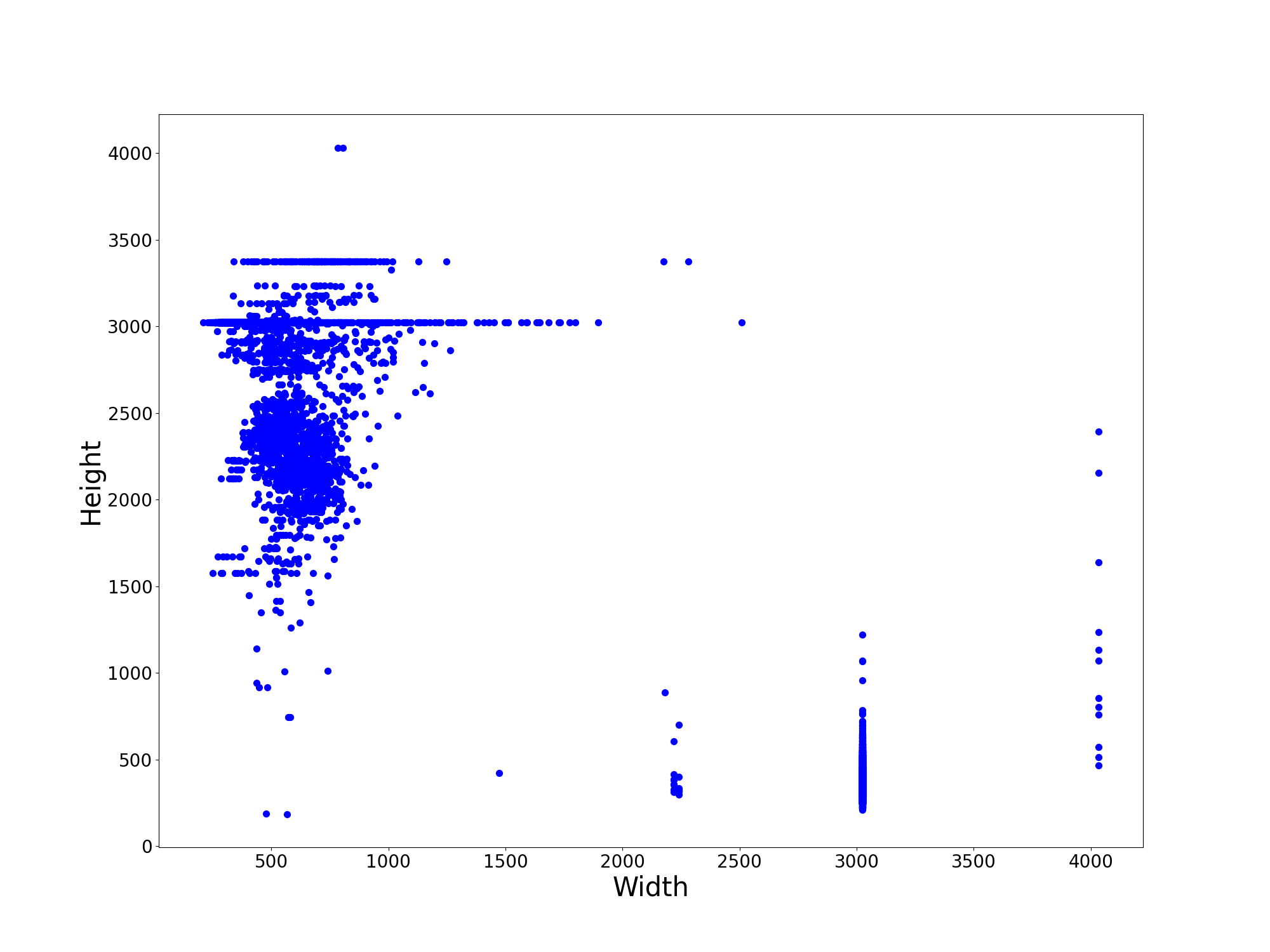}
    \caption{The distribution of image sizes on the pork dataset.}
    \label{fig:size_dist}
\end{figure}

%% file: figure_tex/fusion.tex
\begin{figure}[t]
    \centering
    \includegraphics[width=0.5\textwidth]{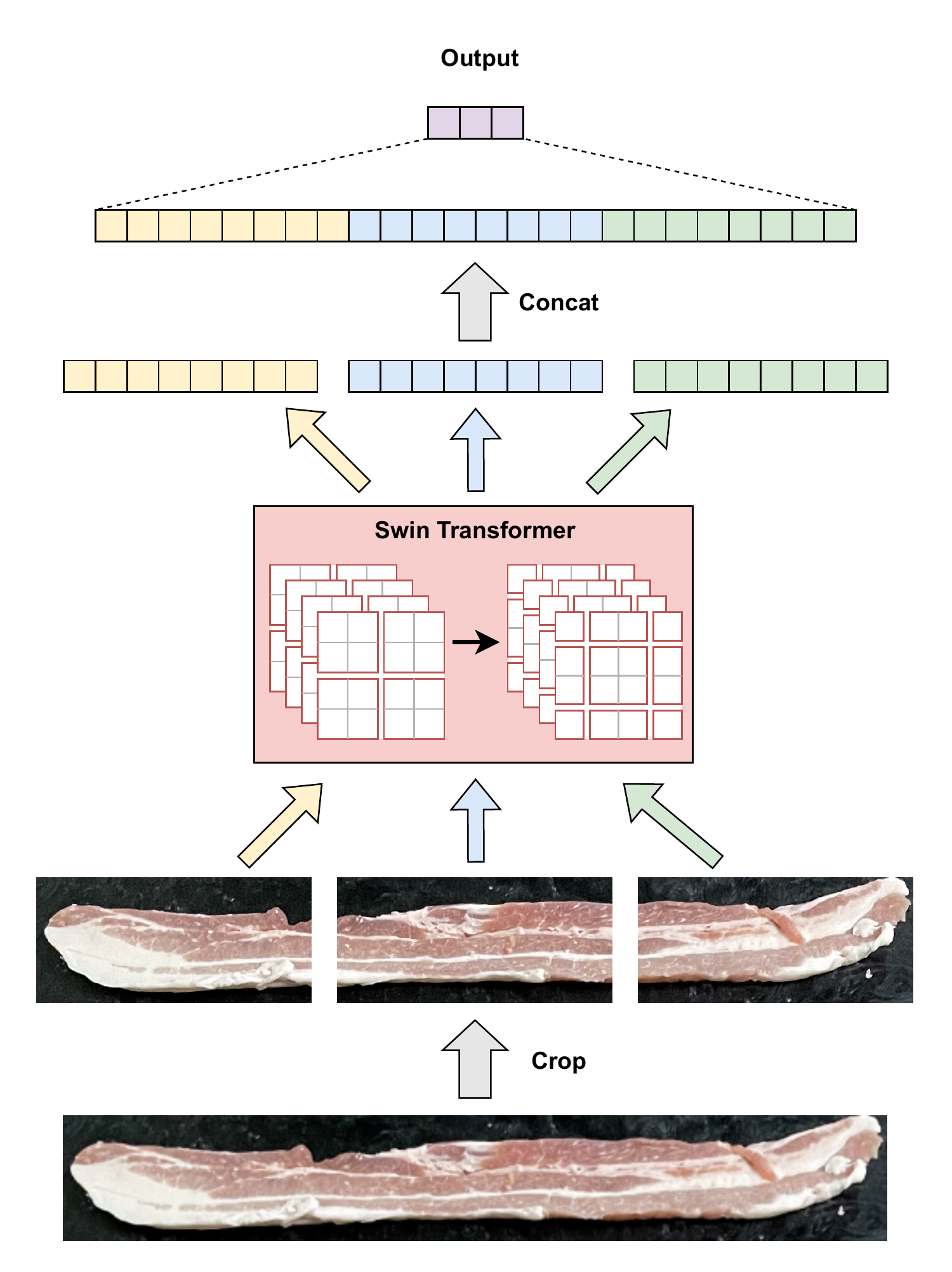}
    \caption{An overview of the multi-view fusion model used for the pork dataset. We simply concatenate the feature vectors of different views before passing through the classification head.}
    \label{fig:fusion}
\end{figure}

%% file: 5_datasets.tex
\input{table_tex/table_total_result}

\section{Datasets} \label{sec:datasets}
We formulate RPQD as a multi-class image classification task where the classes represent quality measures pre-defined for each kind of raw produce.
We train and evaluate on four RPQD datasets from AI Hub, namely fruits and vegetables, pork, beef, and fish.
Datasets vary not only in their domains, but also in their collection strategies and annotation schemes, resulting in diverse sample and label distributions.

\subsection{Fruits \& Vegetables}
Fruits and vegetables dataset consists of roughly 160k images of 10 most consumed domestic fresh produces with high trade volumes in South Korea. 
The images are taken in a studio setting, resulting in high quality images with little to no noise. 
Moreover, the dataset has roughly uniform class distribution, providing a good baseline for experiments.
Each instance of fruit or vegetable was captured at 5 longitudinal angles (i.e. -90, -45, 0, 45, 90 degrees) and 8 latitudinal angles (i.e. -135, -90, -45, 0, 45, 90, 135, 180 degrees), resulting in the total of 40 image samples per instance.
Each sample has type (e.g. apple, persimmon, radish), variety (e.g. fuji (apple), bansi (persimmon), winter (radish)), and grade (e.g. standard, prime, or prime+) attributes.
For the purpose of our quality detection task, we predict one of the three grade attributes of each image sample.

\subsection{Beef}
Beef dataset contains about 78k images obtained from video recordings of a beef inspection facility.
The images contain bisected dressed carcass captured at different angles from which semantic information such as meat color, marbling can be inferred. 
Each sample metadata is annotated by the domain expert and contains grade label (i.e. 1++, 1+, 1, 2, or 3), segmentation vertices for region of interest, and supplementary information such as breed. 
The class distribution is overall reasonably uniform.

\subsection{Pork}
Pork dataset contains about 10k images of sliced pork belly, the most widely consumed pork part in South Korea. 
Each sample contains a single specimen of pork belly classified into one of three grades, 1+, 1, or 2, which is usually determined by the thickness of fat. 
Each annotation also contains a set of segmentation vertices for region of interest (e.g. region by which the grade is determined).
We note that the samples are cropped from a larger image to match the shape of the meat, resulting in an elongated shape with excessive amount of background in some samples.
The number of samples in training set only amounts to roughly 3.8k, providing a testing ground for low-data regimes.

\subsection{Fish}
Fish dataset consists of 450k images captured from video recordings of five most commonly farmed fish species in South Korea, i.e. red sea bream, rock bream, halibut, black porgy, and rock fish. 
Each sample annotation contains disease type and corresponding bounding box information of the affected region of positive subjects.
We note that the dataset has a severe class imbalance, lacking in positive samples by a factor of thousand. 

%% file: table_tex/table_total_result.tex
\begin{table*}[ht]
    \centering
    \ra{1.1}
    \begin{tabular}{@{} r ccc c cccc @{}} \toprule[1.5pt]
         & \multicolumn{3}{c}{ResNet} & BiT & \multicolumn{4}{c}{Swin} \\ \cmidrule(lr){2-4} \cmidrule(lr){5-5} \cmidrule(lr){6-9} 
        & 50 & 101 & 152 & 101x3 & T & S & B & L \\
        \midrule
        \multicolumn{1}{c}{\# of Params.} & 23.5M  & 42.5M  & 58.1M  & 382M    & 27.5M & 48.8M & 86.7M & 195M \\
        \multicolumn{1}{c}{FLOPS}         & 4.14G  & 7.88G & 11.62G & 71.22G & 4.35G & 8.51G & 15.13G & 34.02G \\
        \multicolumn{1}{c}{Throughput (img / sec)}    & 380.36 & 229.37 & 159.26 & 32.80 & 214.48 & 123.68 & 86.04 & 44.43 \\ \midrule
        \multicolumn{1}{c}{Fruits \& Vegetables} & & & & & & & & \\
        \cmidrule(lr){1-1}
        Apple               & 96.69 & 95.82 & 98.31 & 99.47 & \textbf{99.93} & 99.79 & \textbf{99.93} & 99.79 \\
        Pear                & 99.92 & 99.87 & \textbf{100.0} & \textbf{100.0} & \textbf{100.0} & \textbf{100.0} & \textbf{100.0} & \textbf{100.0} \\
        Persimmon           & 98.95 & 98.74 & 99.45 & 99.99 & \textbf{100.0} & \textbf{100.0} & 99.99 & \textbf{100.0} \\ 
        Mandarine           & 95.56 & 95.37 & 98.10 & 99.30 & 99.73 & \textbf{99.79} & 99.57 & 99.52 \\
        Cabbage             & 96.85 & 96.48 & 98.66 & 99.82 & \textbf{100.0} & \textbf{100.0} & 99.93 & 99.87 \\
        Chinese Cabbage     & 95.06 & 95.47 & 96.55 & 98.48 & \textbf{99.06} & 98.95 & 98.78 & 98.78 \\
        Garlic              & 95.41 & 93.87 & 97.12 & 99.08 & \textbf{99.80} & 99.75 & 99.52 & 99.44 \\
        Onion               & 96.11 & 95.49 & 97.92 & 99.41 & \textbf{99.89} & 99.84 & 99.75 & 99.48 \\
        Potato              & 98.01 & 98.31 & 99.02 & 99.89 & 99.97 & \textbf{100.0} & 99.97 & 99.95 \\
        Radish              & 97.09 & 96.20 & 98.53 & 99.70 & 99.86 & \textbf{99.89} & 99.76 & 99.59 \\ \hdashline\noalign{\vskip 0.5ex}
        Total               & 97.04 & 96.63 & 98.43 & 99.56 & \textbf{99.86} & 99.83 & 99.76 & 99.67 \\
        \midrule
        \multicolumn{1}{c}{Pork} & 62.68 & 63.62 & 62.37 & 65.60 & 82.43 & \textbf{83.06} & \textbf{83.06} & 83.03 \\
        \midrule
        \multicolumn{1}{c}{Beef} & 96.95 & 97.57 & 97.71 & \textbf{98.51} & 98.16 & 98.41 & 97.96 & 98.35 \\ \midrule
        \multicolumn{1}{c}{Fish} & 80.64 & 78.11 & 82.68 & 78.97 & 87.62 & 91.13 & \textbf{91.26} & 89.83 \\ \bottomrule[1.5pt]
    \end{tabular}
    \caption{Evaluation results on test split for different models.
    The highest metrics are bolded.}
    \label{tab:total_results}
\end{table*}

%% file: 6_experimental_results.tex
\section{Experimental Results}
We compare Swin Transformers with the previous state-of-the-art CNNs, ResNet \cite{resnet} and BiT \cite{bit}.
For all models we consider, we fine-tune the ImageNet pre-trained models on the RPQD datasets described in Section \ref{sec:datasets}.
We use AUROC (Area Under the Receiver Operating Characteristic curve) and accuracy as the prediction performance metrics for fish and the other datasets, respectively.
For each dataset of raw produce, we randomly shuffle the dataset, and use 0.4, 0.3, and 0.3 splits of the dataset for training, validation, and testing, respectively.

\subsection{Training Details and Hyperparameters}
For Swin Transformers, we follow the default model architectures suggested in ~\cite{swint} (Swin-T, Swin-S, Swin-B, and Swin-L).
The Swin-T and Swin-S, and Swin-B and Swin-L are pre-trained on ImageNet-1K and ImageNet-22K, respectively.
The parameters of Swin Transformers are optimized by AdamW using cosine learning rate scheduler with 20 warm-up epochs.
We use an initial learning rate of 3.125e-5, a weight decay of 0.05, and a gradient clipping with a max norm of 1.
We also employ the stochastic depth augmentation of 0.2, 0.3, 0.5, and 0.5 degree for Swin-T, Swin-S, Swin-B, and Swin-L, respectively.

For ResNet, we use ResNet-50, 101, and 152 variants that are pre-trained on ImageNet-1K. 
For a fair comparison against the large Transformer models pre-trained on ImageNet-21K, we use BiT ResNetV2-101x3 variant, whose every hidden layer is scaled up by a factor 3, and batch normalization is replaced with group normalization and weight standardization to mitigate the effect of small batch size induced by a large model size.
The parameters are optimized by stochastic gradient descent with momentum constant of 0.9 and an initial learning rate of 1e-3.
We reduce the learning rate by a factor of 0.1 when the validation loss plateaus with patience over 5 epochs.

For both Swin Transformers and the CNNs, we use a batch size of 32, input image resolution of $224^{2}$, and follow the same augmentation strategies of \cite{swint} in training.   

\input{table_tex/fusion_ablation}

\subsection{Main Results}
Table~\ref{tab:total_results} shows the test performances of different models. 
We observe that Swin Transformers are able to achieve better performance overall in all of the datasets except for beef, in which BiT scored higher by 0.1\% than the best performance among the Swin Transformers. 
In particular, Swin Transformer scores approximately 26.6 \% and 10.4\% higher than its CNN counterpart in the pork and the fish dataset, respectively.

\subsection{Effect of Multi-view Fusion}
We compare the performance between models trained with and without multi-view fusion in the pork dataset. 
According to Table~\ref{tab:ablation_results}, we observe a significant performance improvement in CNNs, whose performance increased by roughly 26.7\%.
Even though we do not observe as drastic change as in CNNs, Swin Transformers still enjoy about 6.4\% increase with fusion. 

\input{figure_tex/data_efficiency}

\subsection{Robustness to Data-scarcity}
Creating annotations for RPQD requires expert domain knowledge that can come at a prohibitive cost. Hence, data scarcity can be a common problem in actual application.
We decide to investigate the data-efficiency of different models at fine-tuning stage after observing the significant performance gap between CNNs and Swin Transformers in the pork dataset (Table~\ref{tab:total_results}), whose training set only amounts to 3.8k. 
Such performance gap between CNNs and Transformers in scarce data setting has also been observed in \cite{reedha2021vision}.

For this, we create smaller training splits from the fruits \& vegetables dataset by taking half of the samples in the immediately larger training set so that each smaller fraction is a strict subset of the larger fractions. 
From Figure~\ref{fig:data_efficiency} and Table~\ref{tab:data_scarce_results}, we observe that Swin Transformer is highly data-efficient in the low data regimes.
To our surprise, Swin Transformer can achieve 92\% accuracy despite only 1/16-th of the original training samples are used.
On the other hand, CNNs generally performed worse as the model size increased, suggesting that large CNNs are generally less data-efficient.



\input{table_tex/data_efficiency}

\subsection{Trade-off between Prediction Performance and Computation}
For real-world deployment of RPQD models, we should not only consider prediction accuracy but also the computation cost of the models~\cite{rasmussen2020evaluation}.  
Therefore, we take the number of parameters, FLOPS (FLoating point Operations Per Second), and throughputs of the model into account for comparing the memory usage and inference speed. 
We observe that Swin Transformer shows a better accuracy tradeoff in terms of number of parameters, FLOPS, and throughput than CNN, making it ideal for real-world deployment as visualized in Figure~\ref{fig:trade_off}.

\input{figure_tex/tradeoff}
\input{figure_tex/grad_cam}

\subsection{Visual Explanations for Quality Detection}
We visualize the class activation heatmaps of correctly classified samples by Swin Transformer for each domain in Figure~\ref{fig:grad_cam}.
The heatmaps were generated with GradCAM~\cite{selvaraju2017grad}, a common technique for visualizing saliency maps of deep learning models.
The last layer norm of the last transformer block was targeted for visualization.
For datasets with expert annotated segmentation and bounding box labels, we observe that the region of interest mapped by the label and the saliency heatmap align, suggesting that the model is able to learn class-determining features from the images. 


%% file: table_tex/fusion_ablation.tex
\begin{table}[t]
    \centering
    \ra{1.1}
    \begin{tabular}{@{} r cc cc @{}} \toprule[1.5pt]
        & ResNet & BiT & \multicolumn{2}{c}{Swin} \\ \cmidrule(lr){2-2} \cmidrule(lr){3-3} \cmidrule(lr){4-5}
        & 50 & 101x3 & Swin-S & Swin-B \\
        \midrule
        w/o fusion  & 53.18 & 50.23 & 78.40 & 78.09 \\ 
        w fusion    & 62.68 & 63.62 & 83.06 & 83.06 \\
        \bottomrule[1.5pt]
    \end{tabular}
    \caption{Ablation studies on multi-view fusion on the pork dataset.}
    \label{tab:ablation_results}
\end{table}

%% file: figure_tex/data_efficiency.tex
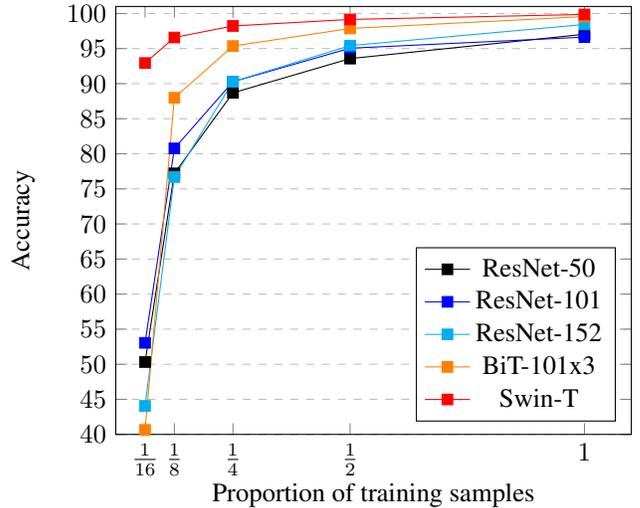
\begin{figure}[t]
\centering
\begin{tikzpicture}
\begin{axis}[
    xlabel={Proportion of training samples},
    ylabel={Accuracy},
    xmin=0, xmax=1.1,
    ymin=40, ymax=101,
    xtick={1/16, 1/8, 1/4, 1/2, 1},
    xticklabels={$\frac{1}{16}$, $\frac{1}{8}$, $\frac{1}{4}$, $\frac{1}{2}$, $1$},
    ytick={40, 45, 50, 55, 60, 65, 70, 75, 80, 85, 90, 95, 100},
    legend pos=south east,
    ymajorgrids=true,
    grid style=dashed,
    ]
\addplot[
    color=black,
    mark=square*,
    ]
    coordinates {
    (1/16, 50.32) (1/8, 77.21) (1/4, 88.69) (1/2, 93.57) (1, 97.04)
    };
    \addlegendentry{ResNet-50}
\addplot[
    color=blue,
    mark=square*,
    ]
    coordinates {
    (1/16, 53.05) (1/8, 80.77) (1/4, 90.27) (1/2, 95.05) (1, 96.63)
    };
    \addlegendentry{ResNet-101}
\addplot[
    color=cyan,
    mark=square*,
    ]
    coordinates {
    (1/16, 44.06) (1/8, 76.69) (1/4, 90.28) (1/2, 95.42) (1, 98.43)
    };
    \addlegendentry{ResNet-152}
\addplot[
    color=orange,
    mark=square*,
    ]
    coordinates {
    (1/16, 40.64) (1/8, 87.99) (1/4, 95.36) (1/2, 97.86) (1, 99.56)
    };
    \addlegendentry{BiT-101x3}
\addplot[
    color=red,
    mark=square*,
    ]
    coordinates {
    (1/16, 92.94) (1/8, 96.58) (1/4, 98.22) (1/2, 99.15) (1, 99.86)
    };
    \addlegendentry{Swin-T}
\end{axis}
\end{tikzpicture}
\caption{Performance comparison between model accuracy and training sample proportions on fruits \& vegetable dataset.}
\label{fig:data_efficiency}
\end{figure}

%% file: table_tex/data_efficiency.tex
\begin{table}[t]
    \centering
    \ra{1.1}
    \begin{tabular}{@{} r ccc c c @{}} \toprule[1.5pt]
        & \multicolumn{3}{c}{ResNet} & BiT & {Swin} \\ \cmidrule(lr){2-4} \cmidrule(lr){5-5} \cmidrule(lr){6-6}
        & 50 & 101 & 152 & 101x3 & Swin-T \\
        \midrule
        \multicolumn{1}{c}{Train Prop.} & & & & & \\
        \cmidrule(lr){1-1}
        1                      & 97.04 & 96.63 & 98.43 & 99.56 & \textbf{99.86} \\
        1/2                    & 93.57 & 95.05 & 95.42 & 97.86 & \textbf{99.15} \\
        1/4                    & 88.69 & 90.27 & 90.28 & 95.36 & \textbf{98.22} \\
        1/8                    & 77.21 & 80.77 & 76.69 & 87.99 & \textbf{96.58} \\
        1/16                   & 50.32 & 53.05 & 44.06 & 40.64 & \textbf{92.94} \\
        \bottomrule[1.5pt]
    \end{tabular}
    \caption{Evaluation results on test split after training on smaller proportions of train split of fruits \& vegetables dataset.}
    \label{tab:data_scarce_results}
\end{table}

%% file: figure_tex/tradeoff.tex
\begin{figure*}[t]
\begin{subfigure}{0.333\textwidth}
\centering
\resizebox{1\textwidth}{!}{
\begin{tikzpicture}
\begin{axis}[
    xlabel={Number of parameters (M)},
    ylabel={Accuracy},
    xmin=10, xmax=310,
    ymin=96, ymax=100,
    xtick={50, 100, 150, 200, 250, 300},
    xticklabels={50, 100, \quad, 200, \quad, 400},
    ytick={96, 97, 98, 99, 100},
    legend pos=south east,
    xmajorgrids=true,
    ymajorgrids=true,
    grid style=dashed,
    ]
\addplot[
    color=blue,
    mark=square*,
    only marks
    ]
    coordinates {
    (23.5, 97.04) (42.5, 96.63) (58.1, 98.43) (290, 99.56)
    };
    \addlegendentry{CNN}
\addplot[
    color=red,
    mark=square*,
    only marks
    ]
    coordinates {
    (27.5, 99.86) (48.8, 99.83) (86.7, 99.76) (195, 99.67)
    };
    \addlegendentry{Swin Transformer}
\end{axis}
\end{tikzpicture}
}
\label{fig:trade_off_params_f&v}
\end{subfigure}
\begin{subfigure}{0.333\textwidth}
\centering
\resizebox{1\textwidth}{!}{
\begin{tikzpicture}
\begin{axis}[
    xlabel={FLOPS (G)},
    ylabel={Accuracy},
    xmin=0, xmax=80,
    ymin=96, ymax=100,
    xtick={0, 20, 40, 60, 80},
    xticklabels={0, 20, 40, 60, 80},
    ytick={96, 97, 98, 99, 100},
    legend pos=south east,
    xmajorgrids=true,
    ymajorgrids=true,
    grid style=dashed,
    ]
\addplot[
    color=blue,
    mark=square*,
    only marks
    ]
    coordinates {
    (4.14, 97.04) (7.88, 96.63) (11.62, 98.43) (71.22, 99.56)
    };
    \addlegendentry{CNN}
\addplot[
    color=red,
    mark=square*,
    only marks
    ]
    coordinates {
    (4.35, 99.86) (8.51, 99.83) (15.13, 99.76) (34.02, 99.67)
    };
    \addlegendentry{Swin Transformer}
\end{axis}
\end{tikzpicture}
}
\label{fig:trade_off_flops_f&v}
\end{subfigure}
\begin{subfigure}{0.333\textwidth}
\centering
\resizebox{1\textwidth}{!}{
\begin{tikzpicture}
\begin{axis}[
    xlabel={Throughputs (img/sec)},
    ylabel={Accuracy},
    xmin=20, xmax=400,
    ymin=95, ymax=100,
    xtick={40, 100, 200, 300, 400},
    xticklabels={40, 100, 200, 300, 400},
    ytick={96, 97, 98, 99, 100},
    legend pos=south east,
    xmajorgrids=true,
    ymajorgrids=true,
    grid style=dashed,
    ]
\addplot[
    color=blue,
    mark=square*,
    only marks
    ]
    coordinates {
    (32.80, 99.56) (159.26, 98.43) (229.37, 96.63) (380.36, 97.04)
    };
    \addlegendentry{CNN}
\addplot[
    color=red,
    mark=square*,
    only marks
    ]
    coordinates {
    (44.43, 99.67) (86.04, 99.76) (123.68, 99.83) (214.48, 99.86)
    };
    \addlegendentry{Swin Transformer}
\end{axis}
\end{tikzpicture}
}
\label{fig:trade_off_thr_f&v}
\end{subfigure}
\begin{subfigure}{0.333\textwidth}
\centering
\resizebox{1\textwidth}{!}{
\begin{tikzpicture}
\begin{axis}[
    xlabel={Number of parameters (M)},
    ylabel={Accuracy},
    xmin=10, xmax=310,
    ymin=96.5, ymax=98.55,
    xtick={50, 100, 150, 200, 250, 300},
    xticklabels={50, 100, \quad, 200, \quad, 400},
    ytick={96.5, 97, 97.5, 98, 98.5},
    legend pos=south east,
    xmajorgrids=true,
    ymajorgrids=true,
    grid style=dashed,
    ]
\addplot[
    color=blue,
    mark=square*,
    only marks
    ]
    coordinates {
    (23.5, 96.95) (42.5, 97.57) (58.1, 97.71) (290, 98.51)
    };
    \addlegendentry{CNN}
\addplot[
    color=red,
    mark=square*,
    only marks
    ]
    coordinates {
    (27.5, 98.16) (48.8, 98.41) (86.7, 97.96) (195, 98.35)
    };
    \addlegendentry{Swin Transformer}
\end{axis}
\end{tikzpicture}
}
\label{fig:trade_off_params_beef}
\end{subfigure}
\begin{subfigure}{0.333\textwidth}
\centering
\resizebox{1\textwidth}{!}{
\begin{tikzpicture}
\begin{axis}[
    xlabel={FLOPS (G)},
    ylabel={Accuracy},
    xmin=0, xmax=80,
    ymin=96.5, ymax=98.55,
    xtick={0, 20, 40, 60, 80},
    xticklabels={0, 20, 40, 60, 80},
    ytick={96.5, 97, 97.5, 98, 98.5},
    legend pos=south east,
    xmajorgrids=true,
    ymajorgrids=true,
    grid style=dashed,
    ]
\addplot[
    color=blue,
    mark=square*,
    only marks
    ]
    coordinates {
    (4.14, 96.95) (7.88, 97.57) (11.62, 97.71) (71.22, 98.51)
    };
    \addlegendentry{CNN}
\addplot[
    color=red,
    mark=square*,
    only marks
    ]
    coordinates {
    (4.35, 98.16) (8.51, 98.41) (15.13, 97.96) (34.02, 98.35)
    };
    \addlegendentry{Swin Transformer}
\end{axis}
\end{tikzpicture}
}
\label{fig:trade_off_flops_beef}
\end{subfigure}
\begin{subfigure}{0.333\textwidth}
\centering
\resizebox{1\textwidth}{!}{
\begin{tikzpicture}
\begin{axis}[
    xlabel={Throughputs (img/sec)},
    ylabel={Accuracy},
    xmin=20, xmax=400,
    ymin=95, ymax=98.55,
    xtick={40, 100, 200, 300, 400},
    xticklabels={40, 100, 200, 300, 400},
    ytick={96.5, 97, 97.5, 98, 98.5},
    legend pos=south east,
    xmajorgrids=true,
    ymajorgrids=true,
    grid style=dashed,
    ]
\addplot[
    color=blue,
    mark=square*,
    only marks
    ]
    coordinates {
    (32.80, 98.51) (159.26, 97.71) (229.37, 97.57) (380.36, 96.95)
    };
    \addlegendentry{CNN}
\addplot[
    color=red,
    mark=square*,
    only marks
    ]
    coordinates {
    (44.43, 98.35) (86.04, 97.96) (123.68, 98.41) (214.48, 98.16)
    };
    \addlegendentry{Swin Transformer}
\end{axis}
\end{tikzpicture}
}
\label{fig:trade_off_thr_beef}
\end{subfigure}
\caption{Accuracy comparison between CNN and Swin Transformer on fruits \& vegetables (top) and beef (bottom) dataset. The horizontal axis of each plot is number of model parameters (left, in millions), FLOPS (middle, in billions), and throughputs (right, in images per second).}
\label{fig:trade_off}
\end{figure*}
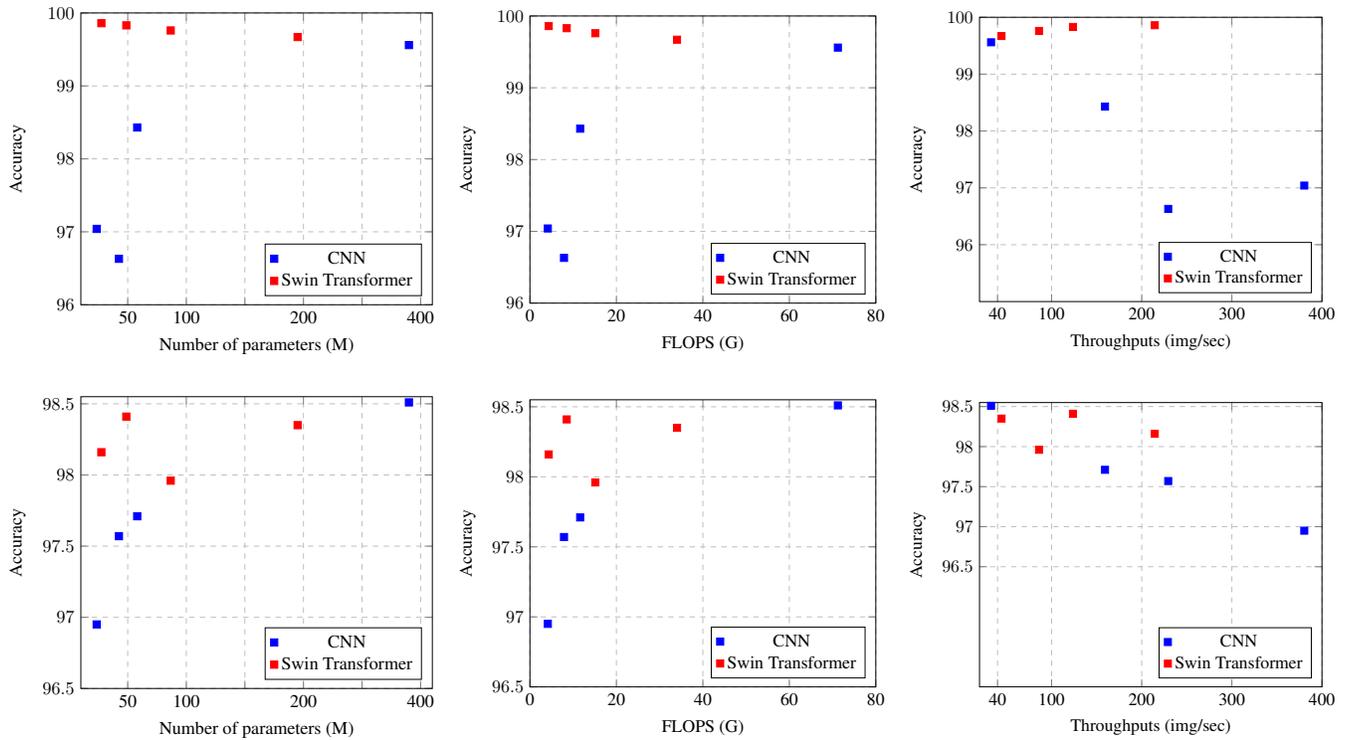

%% file: figure_tex/grad_cam.tex
\begin{figure}[t]
    \centering
    \begin{subfigure}{0.235\textwidth}
        \centering
        \includegraphics[width=\textwidth]{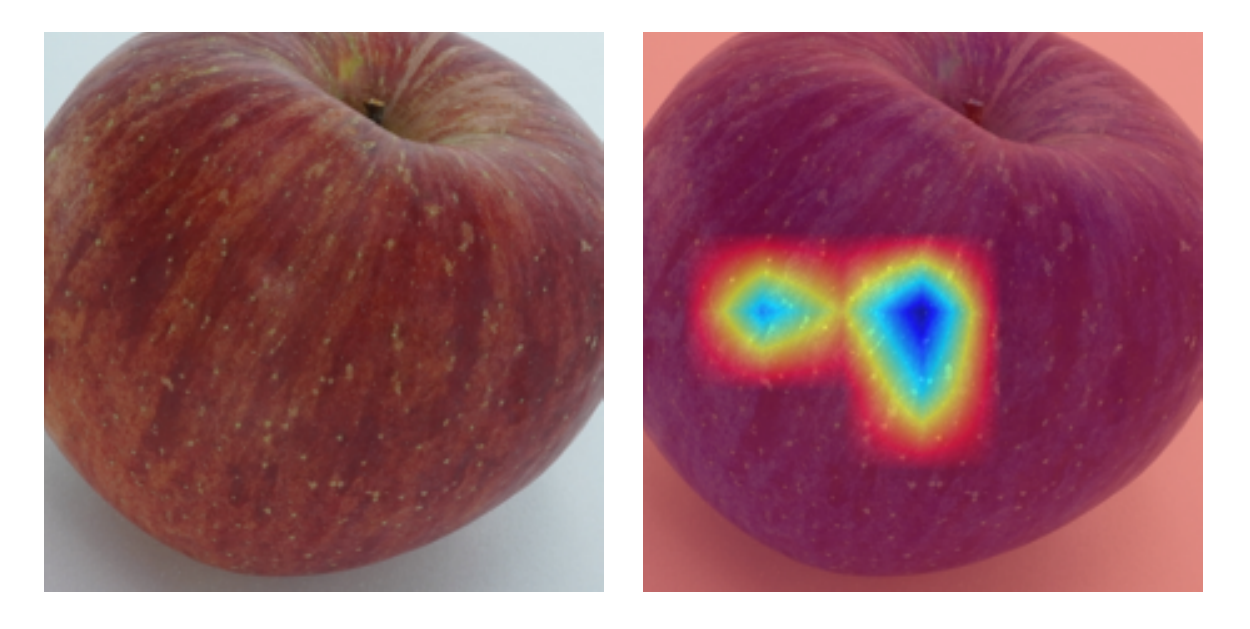}
        \caption{Apple}
        \label{fig:grad_cam_apple}
    \end{subfigure}
    \begin{subfigure}{0.235\textwidth}
        \centering
        \includegraphics[width=\textwidth]{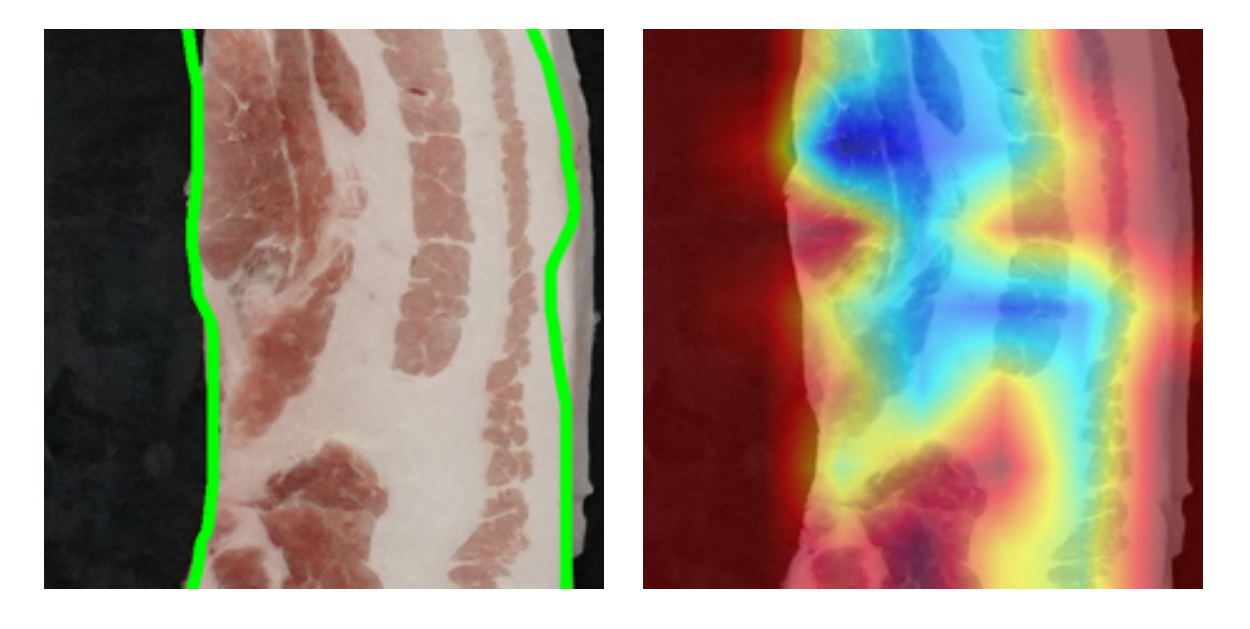}
        \caption{Pork}
        \label{fig:grad_cam_pork}
    \end{subfigure}
    \begin{subfigure}{0.235\textwidth}
        \centering
        \includegraphics[width=\textwidth]{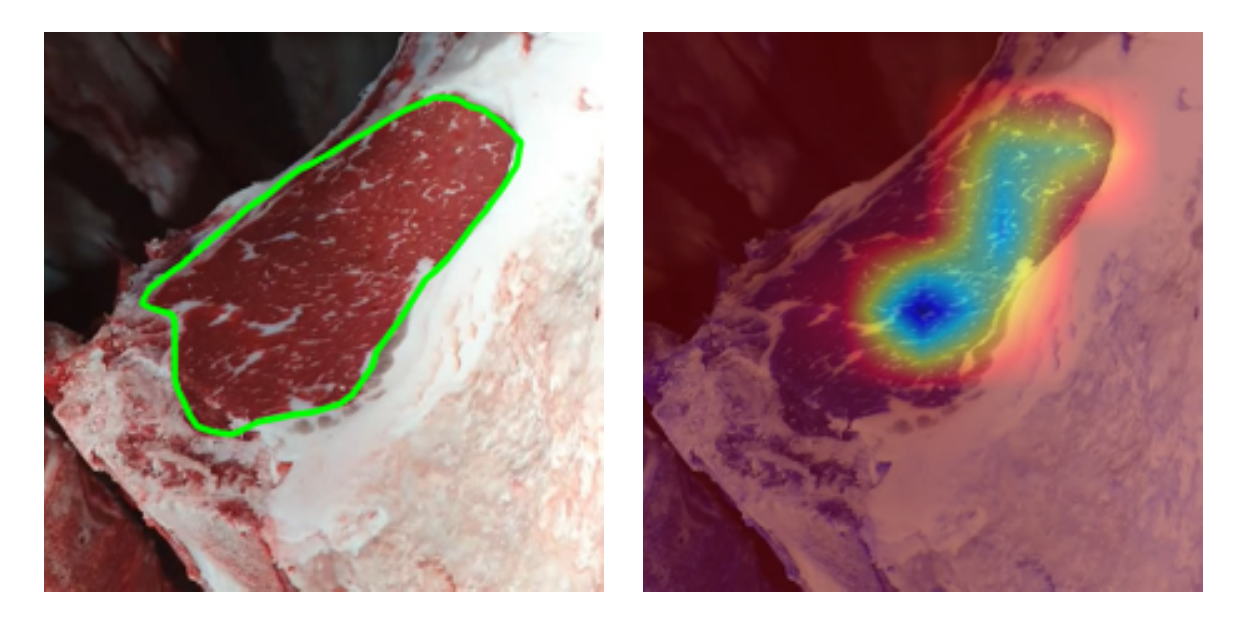}
        \caption{Beef}
        \label{fig:grad_cam_beef}
    \end{subfigure}
    \begin{subfigure}{0.235\textwidth}
        \centering
        \includegraphics[width=\textwidth]{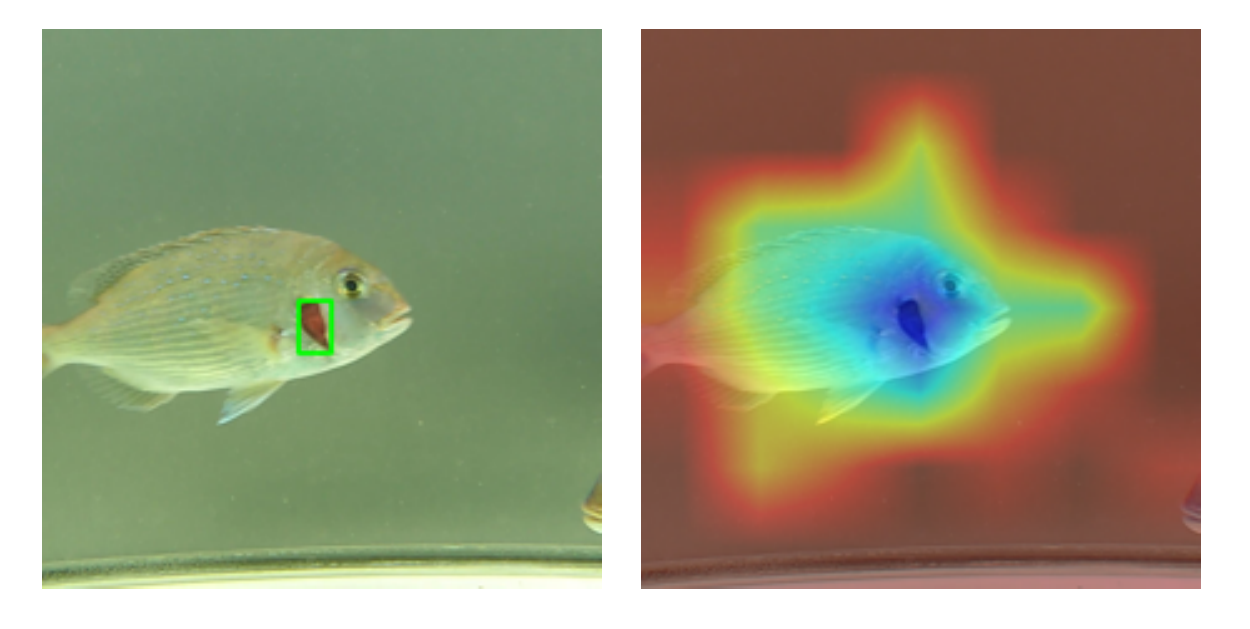}
        \caption{Fish}
        \label{fig:grad_cam_fish}
    \end{subfigure}
    \caption{Class activation visualizations by GradCAM on Swin Transformer. For (b), (c), and (d), we provide ground truth segmentation or bounding box annotations on the left image.}
    \label{fig:grad_cam}
\end{figure}

%% file: 7_conclusion.tex
\section{Conclusion}
In this paper, we compared Swin Transformer against CNN in RPQD task which was formulated as multi-class image classification problem. 
We observed that Swin Transformer performs better than CNN overall, achieving 0.3\%, 26.6\%, and 6.4\% higher in the fruits and vegetables, pork, and fish datasets, respectively, and 0.1\% less in the beef dataset.
Furthermore, we observed the widening performance gap between CNN and Swin Transformer as the number of training samples decreased.
In particular, Swin Transformer was 75.6\% better than the best performing CNN when only 1/16-th of the original training samples were used. 
We also showed that Swin Transformer has a better compute-accuracy tradeoff.
We hope that such practical advantages of Swin Transformer can help improve inspection efficiency in RPQD.
For future work, we hope to see extensive research in RPQD, formulating the task in terms of other vision tasks such as object detection or semantic segmentation.
We also believe multi-view representation learning is essential for the real-world application for RPQD, taking camera angles and metadata into account when designing the models.